\documentclass[conference]{IEEEtran}

\IEEEoverridecommandlockouts
\usepackage{cite}
\usepackage{amsmath,amssymb,amsfonts}
\usepackage{algorithmic}
\usepackage{graphicx}
\usepackage{textcomp}
\usepackage{xcolor}
\usepackage{subcaption}
\def\BibTeX{{\rm B\kern-.05em{\sc i\kern-.025em b}\kern-.08em
		T\kern-.1667em\lower.7ex\hbox{E}\kern-.125emX}}
\usepackage{multirow}

\begin{document}
	
	\title{U-Parking: Distributed UWB-Assisted Autonomous Parking System with Robust Localization and Intelligent Planning\\
		\vspace*{-0.6em}
		\thanks{
			This work was supported in part by Jiangxi Province Science and Technol
			ogy Development Programme under Grant No. 20242BCC32016, in part by
			the National Natural Science Foundation of China under Grant No. 61701197,
			62531015, and U25A20399, in part by the Basic Research Program of
			Jiangsu under Grant BK20252084, in part by the National Key Research and
			Development Program of China under Grant No. 2021YFA1000500(4), in part
			by the Shanghai Kewei under Grant 24DP1500500, in part by the Research
			Grants Council under the Areas of Excellence Scheme under Grant AoE/E
			601/22-R and in part by the 111 Project under Grant No. B23008.
			}
	}
	
	\author{
		\IEEEauthorblockN{
			Yiang Wu\textsuperscript{1,2,3},
			Qiong Wu\textsuperscript{1,2,3,*},
			Pingyi Fan\textsuperscript{4},
			Kezhi Wang\textsuperscript{5},
			Wen Chen\textsuperscript{6},
			Guoqiang Mao\textsuperscript{7},
			Khaled B. Letaief\textsuperscript{8}
		}
		\IEEEauthorblockA{
			\textsuperscript{1}\textit{Jiangnan University}
			\textsuperscript{2}\textit{School of Information Engineering, Jiangxi Provincial Key Laboratory of Advanced Signal}\\
			\textit{Processing and Intelligent Communications, Nanchang University}
			\textsuperscript{3}\textit{Zhuhai Fudan Innovation Institute}\\
			\textsuperscript{4}\textit{Tsinghua University}
			\textsuperscript{5}\textit{Brunel University of London}
			\textsuperscript{6}\textit{Shanghai Jiao Tong University}
			\textsuperscript{7}\textit{Southeast University}\\
			\textsuperscript{8}\textit{The Hong Kong University of Science and Technology}	
			\textsuperscript{*}Cooresponding author is Qiong Wu
			\vspace*{-1em}
		}
	}
	\maketitle
	
	\begin{abstract}
	This demonstration presents U-Parking, a distributed Ultra-Wideband (UWB)-assisted autonomous parking system. By integrating Large Language Models (LLMs)-assisted planning with robust fusion localization and trajectory tracking, it enables reliable automated parking in challenging indoor environments, as validated through real-vehicle demonstrations.
	\end{abstract}
	
	\vspace{-0.5em}
	\section{Introduction}
	\vspace{-0.5em}

		Automatic parking is an important application of intelligent driving. In indoor parking lots, complex layouts, occlusions, and multipath effects make high-precision localization difficult, increasing the challenge of path planning and trajectory tracking and causing parking failures or control oscillations. Existing Simultaneous Localization and Mapping (SLAM)-based memorized parking methods rely on high-precision prior maps and sequential scanning, making them sensitive to environmental changes \cite{b1}. In large-scale facilities, conventional search-based planners also suffer from large search spaces, low efficiency, and limited global decision-making.
		
		Ultra-Wideband (UWB) technology offers centimeter-level accuracy for indoor localization; however, real-world deployments remain susceptible to line-of-sight (LOS) and non-line-of-sight (NLOS) transitions and multipath effects, causing noticeable		positioning jumps that further amplify uncertainties in planning and control. To alleviate these issues, UWB–Inertial Measurement Unit (IMU) fusion is commonly employed to enhance localization robustness \cite{b3}. 
		In parallel, the reasoning capabilities of Large Language Models (LLMs) in complex decision-making tasks introduce new opportunities for high-level planning in autonomous parking \cite{b4}. Recent studies have shown that LLM-guided planning can effectively assist classical search algorithms; for example, reference \cite{b5} demonstrates that LLM-assisted waypoint selection significantly reduces the number of explored nodes in A* search. Inspired by this insight, LLMs can consider parking-lot topology, parking-space availability, and perception reliability to efficiently filter candidate parking spaces, thereby constraining the path search space at the system level and improving planning efficiency.
		
		\vspace{-1.0em}
		\begin{figure}[htbp]
			\centering 
			\includegraphics[width=\linewidth]{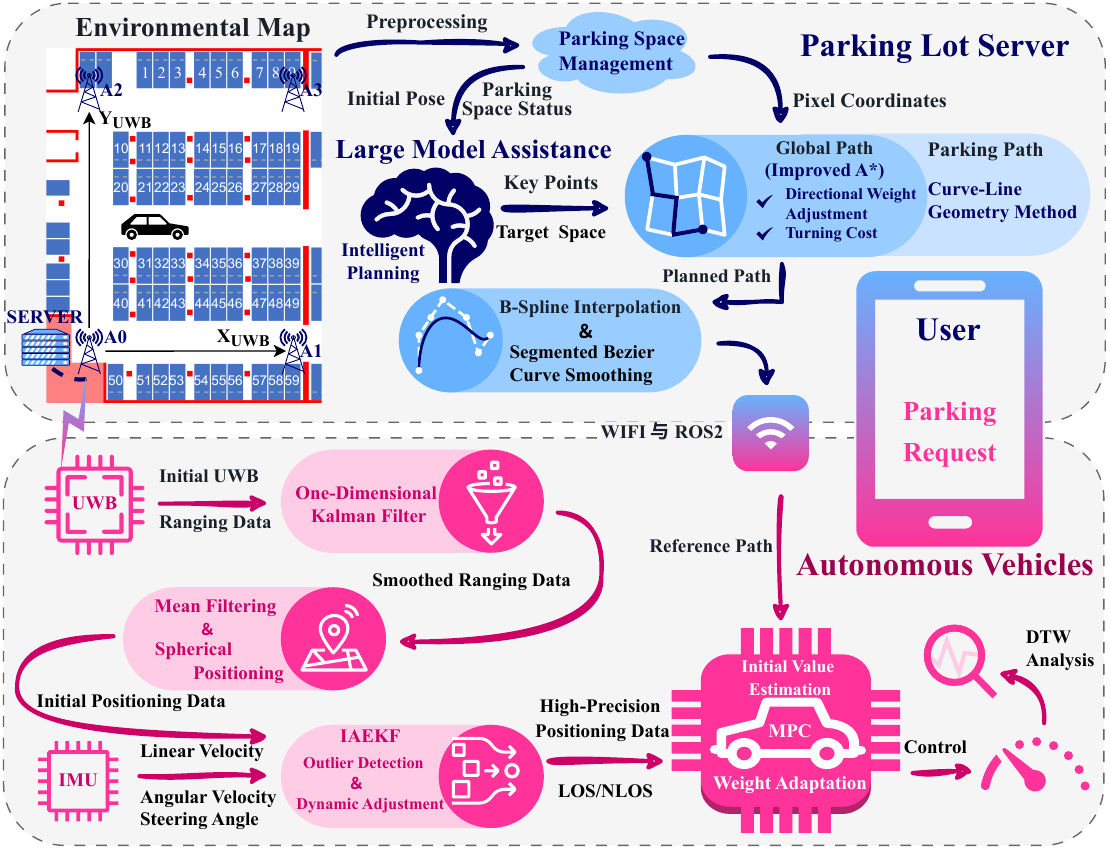}  
			\vspace{-1.5em}
			\caption{The system architecture of U-Parking.}
			\label{fig}
			\vspace{-1.0em}
		\end{figure}	
		
		Based on the above background, as illustrated in Fig.~\ref{fig}, this paper demonstrates a UWB-based vehicle-parking lot collaborative automatic parking system named U-Parking. Built on a distributed Robot Operating System 2 (ROS2) framework, the system introduces an LLM-assisted planning mechanism on the server, combined with layered fusion positioning and robust trajectory tracking control. This enables stable automatic parking under conditions with frequent LOS/NLOS switching.
		
		Compared with existing approaches, U-Parking offers the following advantages: 1) A hierarchical UWB fusion localization strategy that improves robustness during LOS/NLOS transitions, mitigating meter-level drift in conventional UWB systems. 2) 
		An LLM-assisted server-side planning mechanism that provides high-level guidance for parking space selection and path planning, effectively reducing the search space of conventional A* planners.
		 3) A collaborative vehicle–parking lot architecture that enables unified global positioning and decision-making beyond locally perceived maps \cite{b11, b12}.
			
	\vspace{-0.2em}
	\section{Proposed Method}
	\vspace{-0.1em}
	\textbf{Intelligent Parking Space Selection and Path Planning.}
	Unlike traditional methods that rely solely on A* search, U-Parking adopts an LLM-assisted planning framework on the parking-lot server. During system initialization, the parking-lot map is preprocessed to extract obstacle geometry and candidate path nodes \cite{b13, b14}. Upon receiving a parking request, the server invokes the LLM with structured inputs, including obstacle locations, candidate nodes, parking-space IDs and coordinates, UWB anchor deployment, as well as the vehicle’s initial pose and real-time parking-space availability. Based on these inputs, the LLM prioritizes regions with favorable UWB signal quality and consecutive available parking spaces, and outputs a target parking-space ID together with a set of key waypoints \cite{b15, b16}. These waypoints are then used to guide an improved A* planner, significantly reducing unnecessary grid expansions in cluttered environments. The resulting path is further smoothed using piecewise Bézier curves and B-spline interpolation, while the final parking maneuver is generated using a curve–line geometric method to ensure fast and executable parking trajectories.
	
	\textbf{Layered Fusion Positioning Strategy.}
	To address the susceptibility of UWB positioning to NLOS and multipath effects in complex indoor environments, U-Parking employs a three-layer hierarchical fusion localization architecture, with an improved adaptive extended Kalman filter (IAEKF) as the core optimization module.
	The first layer performs data preprocessing by filtering raw UWB ranging measurements to suppress noise.
	The second layer computes an initial two-dimensional position estimate from the filtered ranging data with preliminary smoothing.
	The third layer performs adaptive localization optimization by fusing UWB position estimates with vehicle motion states. Specifically, the innovation residual is defined as $r_k = z_k^{\text{UWB}} - \hat{x}_k^-$, where $z_k^{\text{UWB}}$ is the UWB-derived position measurement and $\hat{x}_k^-$ is the predicted state from IMU propagation.
	When $\|r_k\|$ exceeds an empirically tuned threshold, the filter dynamically reduces the weighting of UWB measurements, thereby improving localization robustness during LOS/NLOS transitions \cite{b17}.
	
	\textbf{Robust Trajectory Tracking Control.}
	Building upon the fusion-based localization, U-Parking adopts an MPC-based trajectory tracking controller designed for positioning uncertainty. The controller adaptively adjusts its cost weights according to localization reliability: when severe UWB-induced jumps are detected, it down-weights instantaneous position errors and penalizes excessive velocity and steering variations to suppress aggressive control and vehicle oscillations.
	
	\vspace{-0.2em}
	\section{Demonstration and Experiment}
	\vspace{-0.2em}
	\textbf{Experimental Setup.}
	Experiments were conducted on Ubuntu 22.04 using ROS2 Humble. The system employs an HR-RTLS1 UWB platform (ULM1 tags and LD150 anchors) covering a 30.2 m × 37.9 m indoor area, a WHEELTEC N100 IMU, and a SCOUT 2.0 differential-drive mobile platform. The LLM-assisted planning module is implemented using the LLaMA-3.3-70B model via a remote API.	
		The source code is available at: https://github.com/qiongwu86/U-Parking.
	
	\begin{figure}[htbp]
		\centering
		\begin{minipage}[t]{0.24\textwidth}
			\centering
			\includegraphics[height=3.9cm,width=\linewidth]{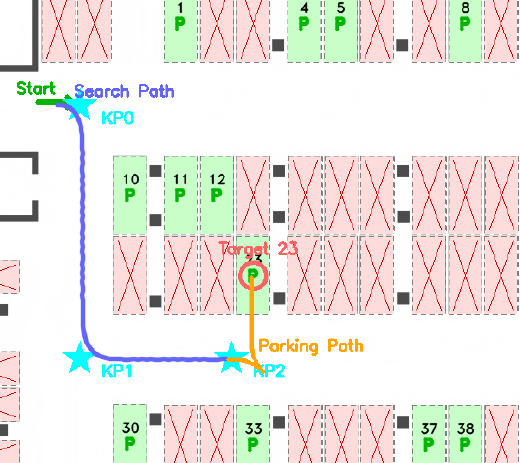}
			\subcaption{Intelligent Planning.}
			\label{fig:location}
		\end{minipage}
		\begin{minipage}[t]{0.24\textwidth}
			\centering
			\includegraphics[height=3.9cm,width=\linewidth]{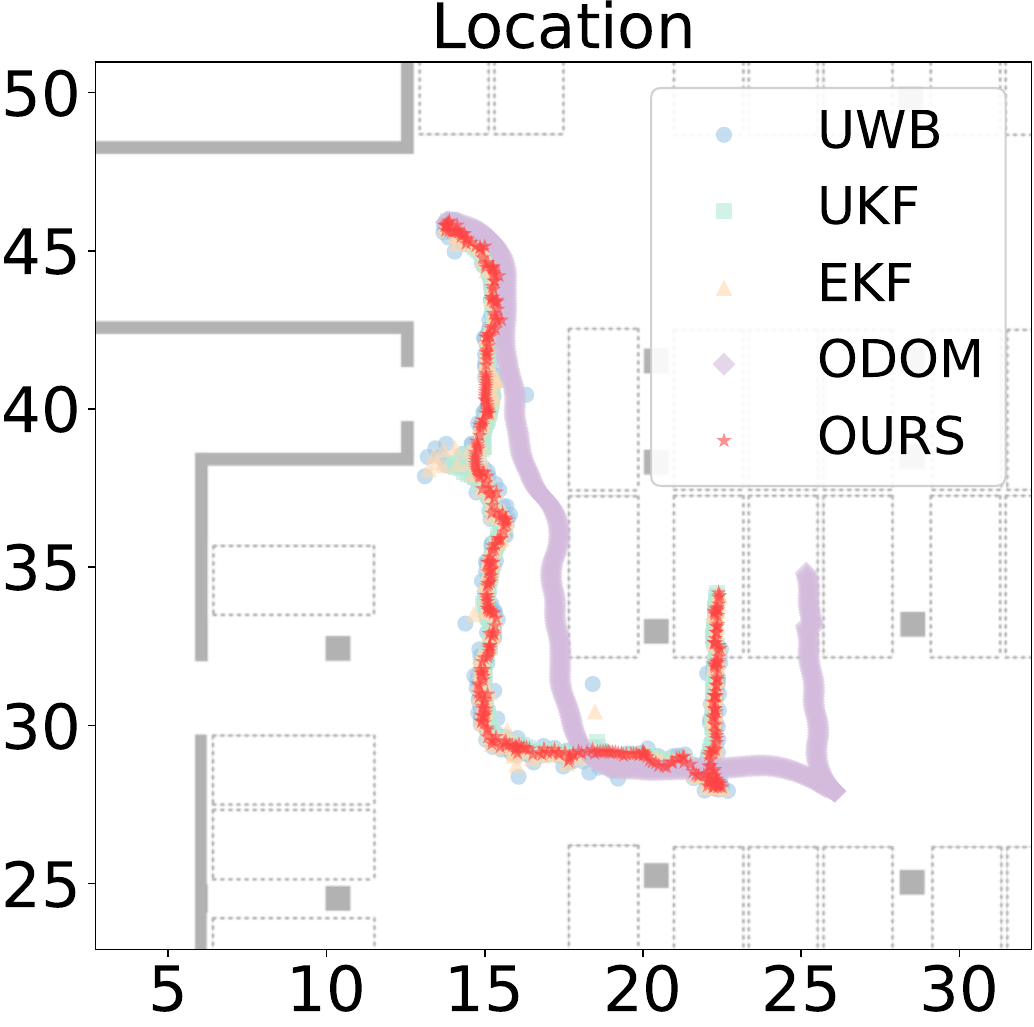}
			\subcaption{UWB Localization Result.}
			\label{fig:parking}
		\end{minipage}
		\vspace{-0.3em}
		\caption{Demonstration.} 
		\label{fig:all_result}
		\vspace{-0.4em}
	\end{figure}
	
	\begin{table}[htbp]
		\caption{Comparison of Tracking Accuracy}
		\vspace{-1em}
		\begin{center}
			\vspace{-0.5em}
			\begin{tabular}{|c|c|c|c|}
				\hline
				\multirow{2}{*}{\textbf{Method}} 
				& \multicolumn{2}{|c|}{\textbf{Euclidean Error (m)}} 
				& \multicolumn{1}{c|}{\textbf{DTW (m)}} \\ 
				\cline{2-4} 
				& Max & Mean & Normalized Mean \\ 
				\hline
				UWB & 2.354 & 0.240 & 0.284 \\ 
				\hline
				UWB+EKF & 1.692 & 0.190 & 0.212 \\ 
				\hline
				UWB+IAEKF & 0.653 & 0.138 & 0.169 \\ 
				\hline
				\parbox{3cm}{\centering \textbf{Integrated}\\\textbf{Improvement (ours)}}
				& \textbf{0.517} & \textbf{0.118} & \textbf{0.133} \\ 
				\hline
			\end{tabular}
			\label{tab:comparison}
		\end{center}
		\vspace{-3em}
	\end{table}
	
	\textbf{Demonstration.}
	U-Parking was implemented and demonstrated on real vehicles in an indoor parking lot. The system operates within a shared local area network, where users submit parking requests and vehicles register with the parking-lot server. As illustrated in Fig. 2(a), the server maintains parking-slot availability, performs intelligent global planning, and assigns smoothed reference trajectories to vehicles for execution.	Each vehicle is equipped with a UWB tag and communicates with a deployed UWB anchor network. Vehicle poses are estimated using the proposed hierarchical fusion localization framework. Based on the planned trajectories and fusion-based localization results, vehicles complete autonomous parking using the proposed robust tracking controller. 
	As shown in Fig. 2(b), the proposed method achieves more stable and accurate localization than baseline approaches. As reported in Table I, trajectory tracking performance is evaluated using Euclidean error and Dynamic Time Warping (DTW). Although occasional peak errors of up to 0.5 m are observed, they are rare and short-lived, primarily caused by UWB measurement noise, and do not affect overall parking smoothness and tracking accuracy.
	\textbf{The live demonstration video is available at:https://youtu.be/44bzJzwfoUU?si=jYpMfv8q8zDdpJTd.}
		
	\vspace{-0.3em}
	\section{Conclusion}
	\vspace{-0.3em}
	This paper demonstrated a practical UWB-assisted autonomous parking system with server–vehicle collaboration. The results show that combining LLM-based high-level planning with robust localization and control effectively improves parking reliability in real indoor environments.

\end{document}